\documentclass{article}

\PassOptionsToPackage{square,numbers}{natbib}
\usepackage[final]{nips_2016}

\usepackage[utf8]{inputenc} 
\usepackage[T1]{fontenc}    
\usepackage{hyperref}       
\usepackage{url}            
\usepackage{booktabs}       
\usepackage{amsfonts}       
\usepackage{nicefrac}       
\usepackage{microtype}      
\usepackage{graphicx} 
\usepackage{amsmath} 
\usepackage{subcaption}

\DeclareMathOperator*{\argmax}{arg\,max}

\title{Factored Contextual Policy Search with \\ Bayesian Optimization}

\author{
  Peter Karkus$^{1,2}$ \And Andras Kupcsik$^{2}$ \And David Hsu$^{1,2}$ \And Wee Sun Lee$^{2}$ \\
  \AND
  \normalfont{$^{1}$NUS Graduate School for Integrative Sciences and Engineering, National University of Singapore}  \\
 $^{2}$School of Computing, National University of Singapore \\
 \texttt{karkus@comp.nus.edu.sg} 
}

\begin{document}

\maketitle

\begin{abstract}
Scarce data is a major challenge to scaling robot learning to truly complex tasks, as we need to generalize locally learned policies over different "contexts". Bayesian optimization approaches to contextual policy search (CPS) offer data-efficient policy learning that generalize over a context space. We propose to improve data-efficiency by factoring typically considered contexts into two components: target-type contexts that correspond to a desired outcome of the learned behavior, e.g. target position for throwing a ball; and environment type contexts that correspond to some state of the environment, e.g. initial ball position or wind speed. Our key observation is that experience can be directly generalized over target-type contexts. Based on that we introduce Factored Contextual Policy Search with Bayesian Optimization for both passive and active learning settings. Preliminary results show faster policy generalization on a simulated toy problem.
\end{abstract}

\section{Introduction}
Robots and other artificial agents act in complex environments where reinforcement learning (RL) is commonly used to learn desired behaviors by interacting with the environment. RL, however, is limited by two main factors: learned policies are difficult to generalize over different problem settings; and they require large amount of data that is often expensive to obtain, due to various costs of interaction with the environment.
In order to perform truly complex tasks we need learning algorithms that generalize well over a context space with little data. 

Factorization is generally used in computer science to decompose problem structures into components that are easier to address. We introduce a factorization of contexts that reduces data requirements for a class of learning tasks, where the context space exhibits specific properties. 

We focus on the contextual policy search (CPS) problem, a formulation of RL, where the concept of \emph{context} was introduced to generalize over task settings. More specifically, we use a Bayesian optimization approach to CPS based on previous work of Metzen et al. \cite{Metzen2015, Metzen2015a}. A key assumption of CPS approaches is that the optimal policy parameters are similar for similar contexts. We observe however, that it is possible to directly generalize experience with some policy parameters over specific context types, which ultimately reduces data requirements.


In the CPS problem the agent is learning an upper level policy $\pi(\theta | s)$ that defines a distribution over the parameter vectors $\theta$ of a lower-level policy for context $s$. Executing the lower level policy with parameters $\theta$ in context $s$ results in an observed trajectory $\tau(s, \theta)$ and reward $R(s, \tau(s, \theta))$.\footnote{We assume that the reward function is fully known or that it can be evaluated for arbitrary $s^t$ targets.} 

The context $s = \{s^e, s^t\}$ may encode two types of task parameters,
\begin{itemize}
\item desired target of the behavior, $s^t$, that only affects how we evaluate a trajectory, $R(s^e, s^t, \tau(s^e, \theta))$, e.g. for a ball throwing task the target coordinates of the ball
\item properties of the environment $s^e$ that influence the trajectory, $\tau(s^e, \theta)$, e.g. initial position of the ball, initial state of the robot arm or wind speed.
\end{itemize}
The key insight of this work is that a single trial provides useful information for \textit{all} target-type contexts. 

Consider a robot learning to throw balls at different targets. First it is asked to aim at target $s^t_1$. It chooses $\theta_1$ according to the upper level policy, execute the throw and observes that the ball hits target $s^t_2 \neq s^t_1$. Even if the trial is poor and results in a low reward the agent gains valuable information: in the future when it is asked to aim at $s^t_2$ it knows that $\theta_1$ parameters will result in hitting $s^t_2$ and yield a high reward.\footnote{Note that the argument above does not hold for environment-type contexts, such as wind speed: an imperfect throw at wind speed $w_1$ does not tell anything about the optimal parameters for any other wind speed $w_2$ apart from the assumed correlation.} Previous work on CPS relies on reward values and an assumed correlation to generalize over contexts. The proposed factorization enables each trial to be evaluated for all target-type contexts leading to faster generalization.  

We introduce Factored Contextual Policy Search with Bayesian Optimization (FCPS-BO), a variant of BO-CPS \cite{Metzen2015} with factored contexts. We also extend FCPS-BO to the active learning setting based on ACES \cite{Metzen2015a}. In active learning the agent may choose the context in each episode during training, unlike the standard (passive) formulation of CPS, where the context is given.


\section{Related work}

There are a number of CPS approaches that generalize over a discrete or continuous context space. One group of work interpolates policy parameters over contexts given a number of local policies \cite{DaSilva2012,Metzen2014}. This approach is limited to problems where local policies are available or easy to learn but inefficient if learning local policies are challenging themselves. 

The second group of work jointly learns local policies and generalizes over the context space.  Reward-Weighted Regression (RWR) \cite{Peters2007} - and its kernelized variant Cost-Regularized Regression (CrKR) \cite{Wilhelm2014} - use a linear function to represent the upper-level policy and perform updates by weighted linear regression. Contextual Relative Entropy Policy Search (C-REPS) \cite{Kupcsik2014} represents the upper-level policy as a probability distribution over the lower-level policy parameters conditioned on the context, while updates are performed using information theoretic insights.
CrKR and C-REPS were applied to a variety of real-world RL tasks: playing table tennis \cite{Wilhelm2014}, throwing darts \cite{Wilhelm2014} and playing hockey \cite{Kupcsik2014}. Although these tasks all involve target-type contexts they are not leveraged to evaluate previous experience in new query contexts.

Recently Metzen et al. introduced BO-CPS \cite{Metzen2015} and ACES \cite{Metzen2015a} that use a Bayesian optimization approach for CPS in passive and active learning settings, respectively. They directly learn a Gaussian Process (GP) model of the expected reward given a parameter vector $\theta$ and context $s$.  These methods offer high data-efficiency and handle noisy observations but are limited to low dimensional parameter spaces and are yet to be evaluated on real world learning tasks. Our method extends these approaches with the concept of factored contexts.

To the best of our knowledge there is no work that explicitly factors the context space. Similar ideas are implicitly used by Kober et al. who apply CrKR to learn a contextual policy for discrete targets while performing a higher-level task \cite{Kober2011}. They map experience gained in one context to another; however, they do so for the purpose of estimating discrete outcome probabilities and not for improving the policy. 

Kupcsik et al. introduced GP-REPS \cite{Kupcsik2014a}, a model-based policy search approach that iteratively learns a transition model of the system using a GP. The GP prediction is then used to generate trajectories offline for updating the policy. The authors discuss the option of generating additional samples for artificial contexts - which is a similar idea to our mapping method, -but  they do not define an explicit factorization and generate multiple samples for each observed trajectory rather than explicitly mapping to the query context. 


\section{Factored Contextual Policy Search with Bayesian Optimization}
The Bayesian optimization approach to CPS trains a GP to directly map from policy parameters and contexts to the expected reward, $E[R(s, \theta)]$, given all previous experience,
$D = \{s_i, \theta_i, R_i\}$.
The GP posterior is an estimate of the reward $\mu[R^*(s, \theta)]$ with uncertainty $\sigma[R^*(s, \theta)]$. 

In each episode the agent is given a query context $s_q$ and needs to choose an appropriate parameter vector $\theta^*$. We use the acquisition function,
\begin{equation}
\textrm{GP-UCB}(s_q, \theta) = \mu[R^*(s_q, \theta)] + \kappa \sigma[R^*(s_q, \theta)],
\end{equation}
where $\kappa$ controls the exploration-exploitation trade-off.
The policy parameters are selected by simply optimizing the acquisition function given the query context, 
\begin{equation}
\theta^* = \argmax_{\theta}{\textrm{GP-UCB}(s_q, \theta)}.
\end{equation}

In order to leverage factored contexts we maintain a dataset $D = \{s^e_i, \theta_i, \tau_i\}$ with trajectories $\tau_i$ observed when executing the policy with $\theta_i$ parameters in a context $s^e_i$.
Note that we store full trajectories instead of rewards.\footnote{In a practical implementation, if $\tau$ is too large or not observed, we may only store a mapping of the trajectory $f(\tau)$ sufficient for computing the reward for arbitrary target-type contexts. E.g. for a ball throwing task we may only need to store the coordinates where the ball hit rather than the full trajectory.} 
At the query stage, given $s_q = \{s^t_q, s^e_q\}$, we evaluate all previous experience in light of the current target-type query context $s^t_q$. We compute the reward for all previous trajectories,
\begin{equation}
R^{q}_i = R^t(s^t_q, s^r_i, \tau_i),
\end{equation}
where $s^r_i$ and $\tau_i$ are recorded data points and $s^t_q$ is the target context for the query point. We form the dataset
$D^q  = \{s^e_i, \theta_i, R^q_i\}$
that is specific to the query context. The optimal $\theta^*$ parameters for context $s_q$ are then found by optimizing the GP prediction given data $D^{q}$.
By evaluating all previous experience for the current query, we directly generalize over target-type contexts instead of solely relying on assumed correlation as is the case for environment-type contexts.

\section{Active learning setting}
In the active learning setting the agent chooses both parameters $\theta$ and context $s$ for the next trial. Optimizing $\textrm{GP-UCB}$ for $\theta$ and $s$ would not lead to a good result as the achievable rewards for different contexts may be different. Instead we follow ACES \cite{Metzen2015a} and use an Entropy Search based acquisition function\footnote{Entropy search could be used in the passive setting as well. We plan to do such experiments in the future.} which aims to choose the most informative query points for global optimization \cite{Hennig2012}.

In addition to a GP mapping from contexts and parameters to expected rewards, ACES maintains an explicit probability distribution $p_{opt}(\theta | s)$, the probability of $\theta$ being the global optimal in context $s$. The value of $p_{opt}$ is approximated on a set of heuristically chosen representer points using Monte Carlo integration. ACES predicts $p_{opt}[s_q \theta_q](\theta | s)$, the distribution after a hypothetical query at $\{s_q, \theta_q\}$. To do that we need to predict the GP posterior after a query. The new posterior is approximated by first sampling from the current GP posterior at the query point and then adding the sample to the GP training data.

The most informative query point is chosen by maximizing the change in relative entropy of $p_{opt}$ integrated over the context space. More specifically we define a loss function for a query point at context $s$,
\begin{equation}
L^s(s_q, \theta_q) = L(p_{opt}[s_q \theta_q](\theta | s)) -  L(p_{opt}(\theta | s)),
\end{equation}
where
$L(p_{opt})$ is the negative relative entropy between $p_{opt}$ and a uniform measure.
The acquisition function is then the sum of loss functions over different contexts,
\begin{equation}
\textrm{ACES}(s_q, \theta_q) = \sum\nolimits_{i=1}^N L^{s_i}(s_q, \theta_q),
\end{equation}
where $\{s_i\}_{i=1}^N$ is a set of randomly chosen representer points. The next query point $\{s_q, \theta_q\}$ is selected by minimizing $\textrm{ACES}(s_q, \theta_q)$.
For further details we refer to \cite{Metzen2015a} and \cite{Hennig2012}.

We integrate factored contexts as follows. In each iteration we map previous experience to all $N$ representer points in the context space, $s_i$, resulting in a set of GP models, $\textrm{GP}_i, i=1..N$. When the loss function $L^s(s_q, \theta_q)$ is evaluated for $s=s_i$ we predict the change in $p_{opt}$ using the corresponding $GP_i$. This way we directly generalize over the target-type context space, similarly to the passive learning case.
Note that the choice on the target-type query context $s_q^t$ is indifferent if we ignore rewards during training, and thus we only need to select $\{s_q^e, \theta_q\}$ by minimizing $\textrm{ACES}(s_q^e, \theta_q)$.

\section{Results}

\begin{figure*}[ht]
    \centering
    \begin{subfigure}[t]{0.5\textwidth}
        \centering
        \includegraphics[width=0.9\textwidth]{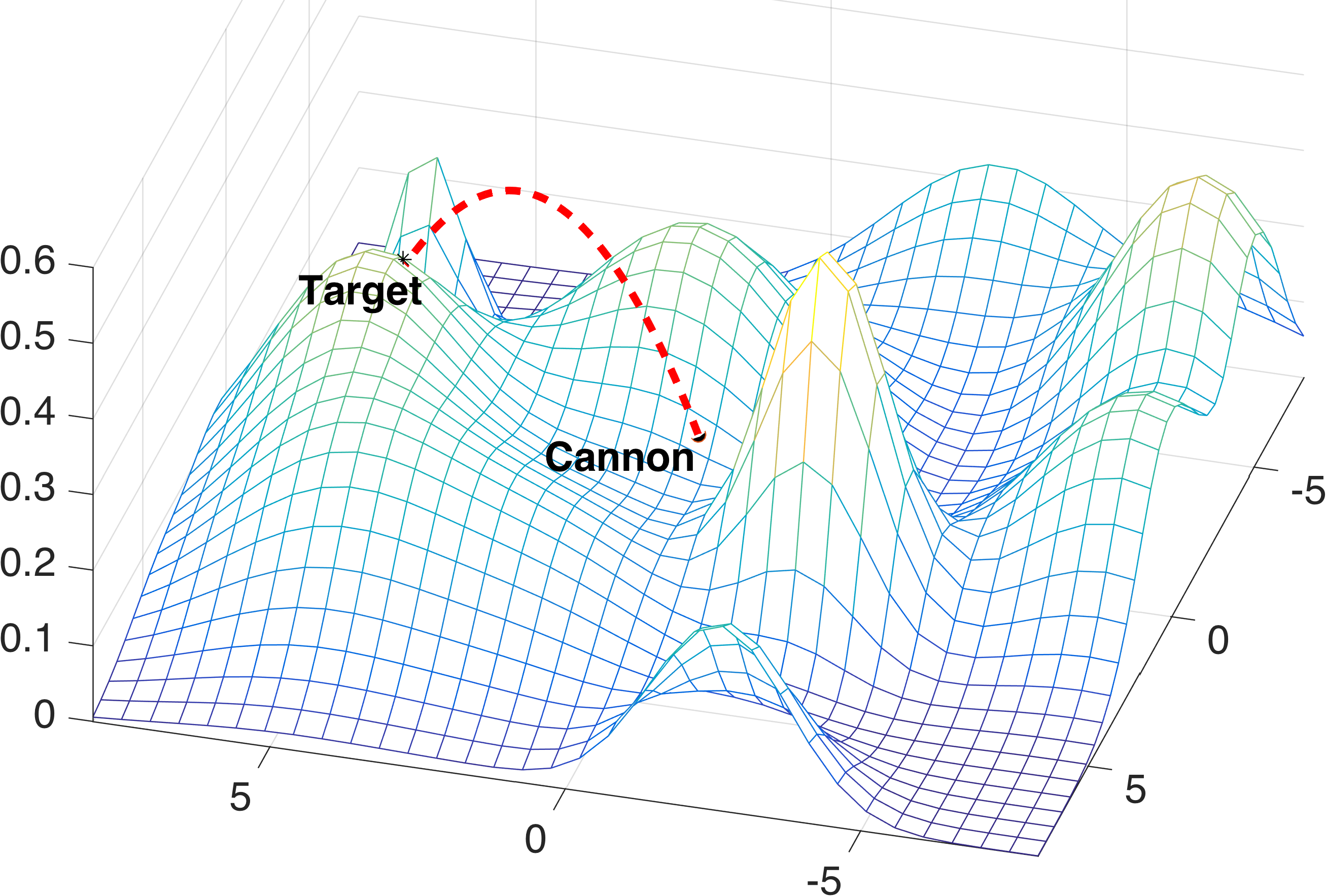}
        \caption{\label{fig:figa}Toy cannon problem}
    \end{subfigure}%
    ~ 
    \begin{subfigure}[t]{0.5\textwidth}
        \centering
        \includegraphics[width=0.86\textwidth]{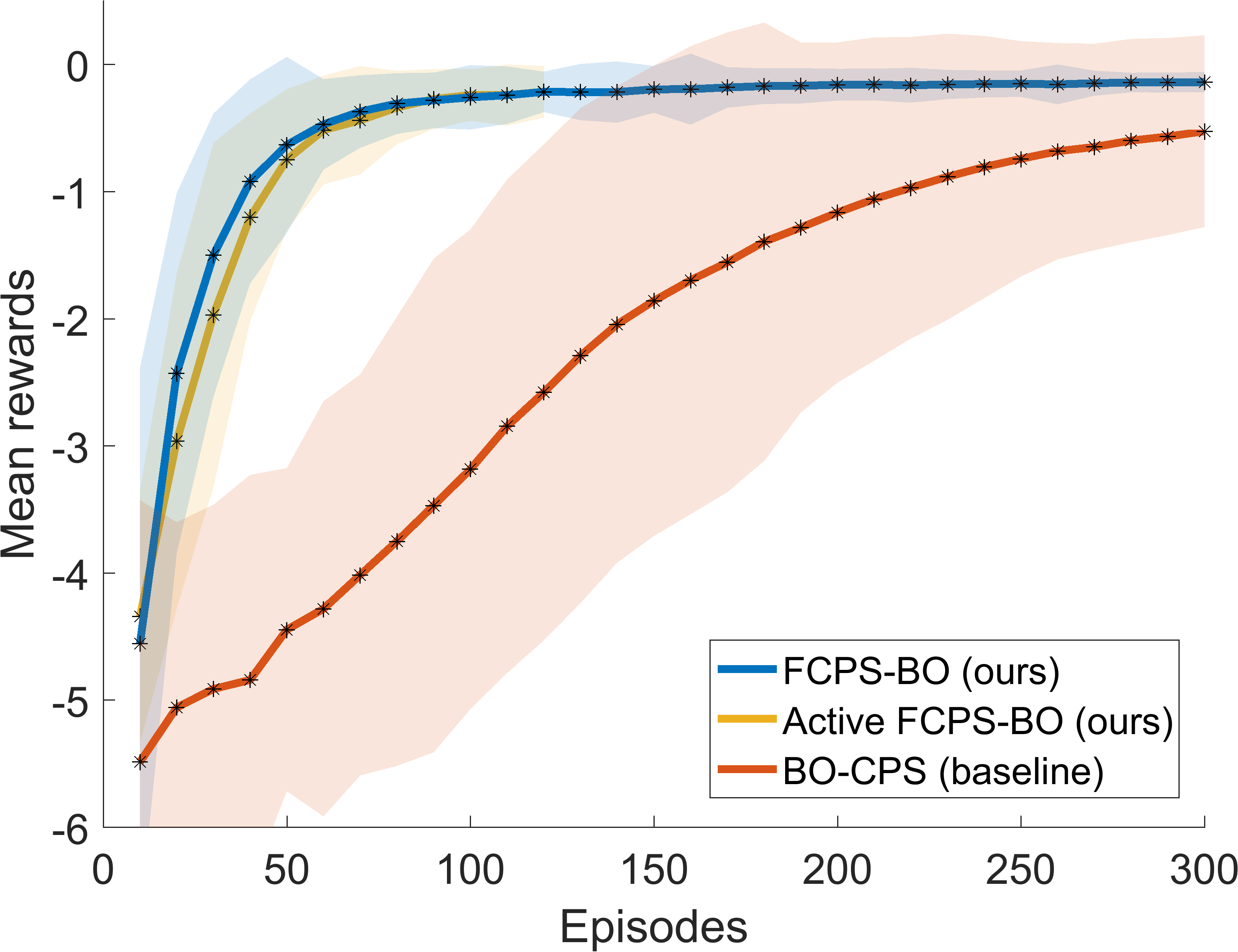}
        \caption{\label{fig:figb} Learning performance}
    \end{subfigure}
    \caption{\label{fig:figure} \subref{fig:figa}) Example for an environment with a target and the optimal trajectory. \subref{fig:figb}) Mean rewards using FCPS-BO (blue), FCPS-BO in active setting (yellow), and the baseline BO-CPS (orange). The shaded area corresponds to two standard deviations.}
\end{figure*}

We run FCPS-BO on a simulated toy cannon problem (Figure \ref{fig:figure}\subref{fig:figa}), where evaluating previous trials for all targets is trivially beneficial. The task is inspired by \cite{DaSilva2014} and is similar in nature to the ball throwing task of \cite{Metzen2015,Metzen2015a}. A cannon is placed in the center of a $3$D coordinate system and has to shoot at targets on the ground in the range of $[-11,11]\times [-11,11]$m. The contextual policy maps from $2$D targets to $3$D launch parameters: horizontal orientation $[0,6.28]$, vertical angle $[0.01, 1.37]$ and velocity $[0.1, 5]\frac{m}{s}$. The reward is a negative quadratic function of the distance between the target and where the projectile hits, plus two terms penalizing larger launch speeds and vertical angles. 

To increase the difficulty of the problem we randomly place hills in the environment. The learning agent is unaware of the hills and the target coordinates carry no information on the target elevation. Furthermore, we add Gaussian noise ($\sigma_n = 1^{\circ}$) to the desired launch angle during training. 

Note that evaluating previous trials for the given context is trivially beneficial in our problem setting. Shooting in a direction opposite to one target results in extremely low reward, but the same launch parameters may achieve high reward when another target is in the shooting direction.

We run trials in $50$ randomly generated environments and evaluate policies offline, with test targets placed on an $8 \times 8$ grid (Figure \ref{fig:figure}\subref{fig:figb}). Our factored approach, FCPS-BO, achieves near optimal performance after less than $100$ episodes, while the baseline BO-CPS requires over $300$ episodes. We also evaluate FCPS-BO in active learning setting; however, we observe no benefit over the passive setting. This may be due to the aggressive approximations used in the active approach; or due to our problem setting, where it is not more difficult to learn a good policy for one target than for another. Further investigation is to be conducted in the future.


\section{Conclusion}
We introduced FCPS-BO, a Bayesian optimization approach to CPS that factors the context space to generalize policies with little amount of data. We present an extension of FCPS-BO to the active learning setting, and show the benefit of factorization on a toy problem. Addressing scalability to higher dimensions and evaluation on more realistic problems remain future work. 
We also plan to apply factorization to standard policy search methods, such as C-REPS and GP-REPS.




\small
\bibliography{mendeley}

\begin{thebibliography}{11}
\providecommand{\natexlab}[1]{#1}
\providecommand{\url}[1]{\texttt{#1}}
\expandafter\ifx\csname urlstyle\endcsname\relax
  \providecommand{\doi}[1]{doi: #1}\else
  \providecommand{\doi}{doi: \begingroup \urlstyle{rm}\Url}\fi

\bibitem[da~Silva et~al.(2012)da~Silva, Konidaris, Barto, and
  Castro]{DaSilva2012}
B.~C. da~Silva, G.~Konidaris, A.~G. Barto, and B.~Castro.
\newblock {Learning Parameterized Skills}.
\newblock \emph{ICML}, pages 1679--1686, 2012.

\bibitem[da~Silva et~al.(2014)da~Silva, Konidaris, and Barto]{DaSilva2014}
B.~C. da~Silva, G.~Konidaris, and A.~Barto.
\newblock {Active Learning of Parameterized Skills}.
\newblock \emph{ICML}, 32:\penalty0 1737--1745, 2014.

\bibitem[Hennig and Schuler(2012)]{Hennig2012}
P.~Hennig and C.~J. Schuler.
\newblock {Entropy Search for Information-Efficient Global Optimization}.
\newblock \emph{Machine Learning Research}, 13:\penalty0 1809--1837, 2012.

\bibitem[Kober and Peters(2011)]{Kober2011}
J.~Kober and J.~Peters.
\newblock {Learning elementary movements jointly with a higher level task}.
\newblock \emph{IEEE International Conference on Intelligent Robots and
  Systems}, pages 338--343, 2011.

\bibitem[Kober et~al.(2012)Kober, Wilhelm, Oztop, and Peters]{Wilhelm2014}
J.~Kober, A.~Wilhelm, E.~Oztop, and J.~Peters.
\newblock {Reinforcement learning to adjust parametrized motor primitives to
  new situations}.
\newblock \emph{Autonomous Robots}, 33\penalty0 (4):\penalty0 361--379, 2012.

\bibitem[Kupcsik et~al.(2013)Kupcsik, Deisenroth, Peters, Loh, Vadakkepat, and
  Neumann]{Kupcsik2014}
A.~Kupcsik, M.~P. Deisenroth, J.~Peters, A.~P. Loh, P.~Vadakkepat, and
  G.~Neumann.
\newblock {Data-Efficient Generalization of Robot Skills with Contextual Policy
  Search}.
\newblock \emph{Proceedings of the Twenty-Seventh AAAI Conference on Artificial
  Intelligence}, pages 1401--1407, 2013.

\bibitem[Kupcsik et~al.(2014)Kupcsik, Deisenroth, Peters, Loh, Vadakkepat, and
  Neumann]{Kupcsik2014a}
A.~Kupcsik, M.~P. Deisenroth, J.~Peters, A.~P. Loh, P.~Vadakkepat, and
  G.~Neumann.
\newblock {Model-based contextual policy search for data-efficient
  generalization of robot skills}.
\newblock \emph{Artificial Intelligence}, 2014.

\bibitem[Metzen(2015)]{Metzen2015a}
J.~H. Metzen.
\newblock {Active Contextual Entropy Search}.
\newblock \emph{arXiv preprint arXiv:1511.04211}, 2015.

\bibitem[Metzen et~al.(2014)Metzen, Fabisch, Senger, {de Gea Fernandez}, and
  Kirchner]{Metzen2014}
J.~H. Metzen, A.~Fabisch, L.~Senger, J.~{de Gea Fernandez}, and E.~A. Kirchner.
\newblock {Towards Learning of Generic Skills for Robotic Manipulation}.
\newblock \emph{KI-K{\"{u}}nstliche Intelligenz}, 28:\penalty0 15--20, 2014.

\bibitem[Metzen et~al.(2015)Metzen, Fabisch, and Hansen]{Metzen2015}
J.~H. Metzen, A.~Fabisch, and J.~Hansen.
\newblock {Bayesian Optimization for Contextual Policy Search}.
\newblock \emph{Proceedings of the Second Machine Learning in Planning and
  Control of Robot Motion Workshop, Hamburg}, 2015.

\bibitem[Peters and Schaal(2007)]{Peters2007}
J.~Peters and S.~Schaal.
\newblock {Applying the episodic natural actor-critic architecture to motor
  primitive learning}.
\newblock \emph{ESANN}, pages 295--300, 2007.

\end{thebibliography}
\bibliographystyle{abbrvnat}


\end{document}